%% file: _main.tex
\documentclass[10pt,twocolumn,letterpaper]{article}
\usepackage[most]{tcolorbox} 
\usepackage{wacv}              

\input{preamble}

\definecolor{wacvblue}{rgb}{0.21,0.49,0.74}
\usepackage[pagebackref,breaklinks,colorlinks,allcolors=wacvblue]{hyperref}
\usepackage{epstopdf}
\DeclareGraphicsExtensions{.png,.pdf}
\def\wacvPaperID{1642} 
\def\confName{WACV}
\def\confYear{2026}
\usepackage{comment}

\newcommand\blfootnote[2]{%
  \begingroup
    \renewcommand\thefootnote{}
    \footnote{#1}%
    \addtocounter{footnote}{-1}
  \endgroup
}

\title{Explaining the Unseen: Multimodal Vision-Language Reasoning for Situational Awareness in Underground Mining Disasters}

\author{
Mizanur Rahman Jewel$^{1}$,
Mohamed Elmahallawy$^{2}$,
Sanjay Madria$^{1}$,
Samuel Frimpong$^{1}$ \\
$^{1}$Missouri University of Science and Technology,~
$^{2}$Washington State University \\
\texttt{\{mj9vc, madrias, frimpong\}@mst.edu}, \quad
\texttt{mohamed.elmahallawy@wsu.edu}
}

\usepackage{xcolor}     
 
\usepackage{atbegshi}  

\makeatletter

\newcommand{\FirstPageBannerText}{%
  \parbox{\textwidth}{\large \underline{
    Appears at The IEEE/CVF Winter Conference on Applications of Computer Vision (WACV 2026)}}%
}

\AtBeginShipoutFirst{%
      \raisebox{-0.5\baselineskip}{\FirstPageBannerText}
}

\makeatother
\begin{document}
 
\maketitle

\blfootnote{This research was supported by a grant from CDC--NIOSH.}

\input{sec/0_abstract}

\input{sec/1_intro}
\input{sec/2_related_work}

\input{sec/3_methodology}
\input{sec/4_experiment}

\input{sec/6_conclusion}

\clearpage
{
    \small
    \bibliographystyle{ieeenat_fullname}
    \bibliography{main}
}

\end{document}

%% file: preamble.tex
\usepackage{graphicx} 
\newcommand*{\Resize}[2]{\resizebox{#1}{!}{$#2$}}%


%% file: sec/0_abstract.tex
\begin{abstract}
Underground mining disasters produce pervasive darkness, dust, and collapses that obscure vision and make situational awareness difficult for humans and conventional systems. To address this, we propose \emph{MDSE}, Multimodal Disaster Situation Explainer, a novel vision-language framework that automatically generates detailed textual explanations of post-disaster underground scenes. MDSE has three-fold innovations: (i) \textit{Context-Aware Cross-Attention} for robust alignment of visual and textual features even under severe degradation; (ii) \textit{Segmentation-aware dual pathway visual encoding} that fuses global and region-specific embeddings; and (iii) \textit{Resource-Efficient Transformer-Based Language Model} for expressive caption generation with minimal compute cost. To support this task, we present the Underground Mine Disaster (UMD) dataset—the first image–caption corpus of real underground disaster scenes—enabling rigorous training and evaluation. Extensive experiments on UMD and related benchmarks show that MDSE substantially outperforms state-of-the-art captioning models, producing more accurate and contextually relevant descriptions that capture crucial details in obscured environments, improving situational awareness for underground emergency response. The code is at \href{https://github.com/mizanJewel/Multimodal-Disaster-Situation-Explainer}{Github}.
\end{abstract}

%% file: sec/1_intro.tex
\section{Introduction}\label{sec:intro}

\begin{figure}[!t] 
    \centering 
    \includegraphics[width=0.8\linewidth, height=5cm]{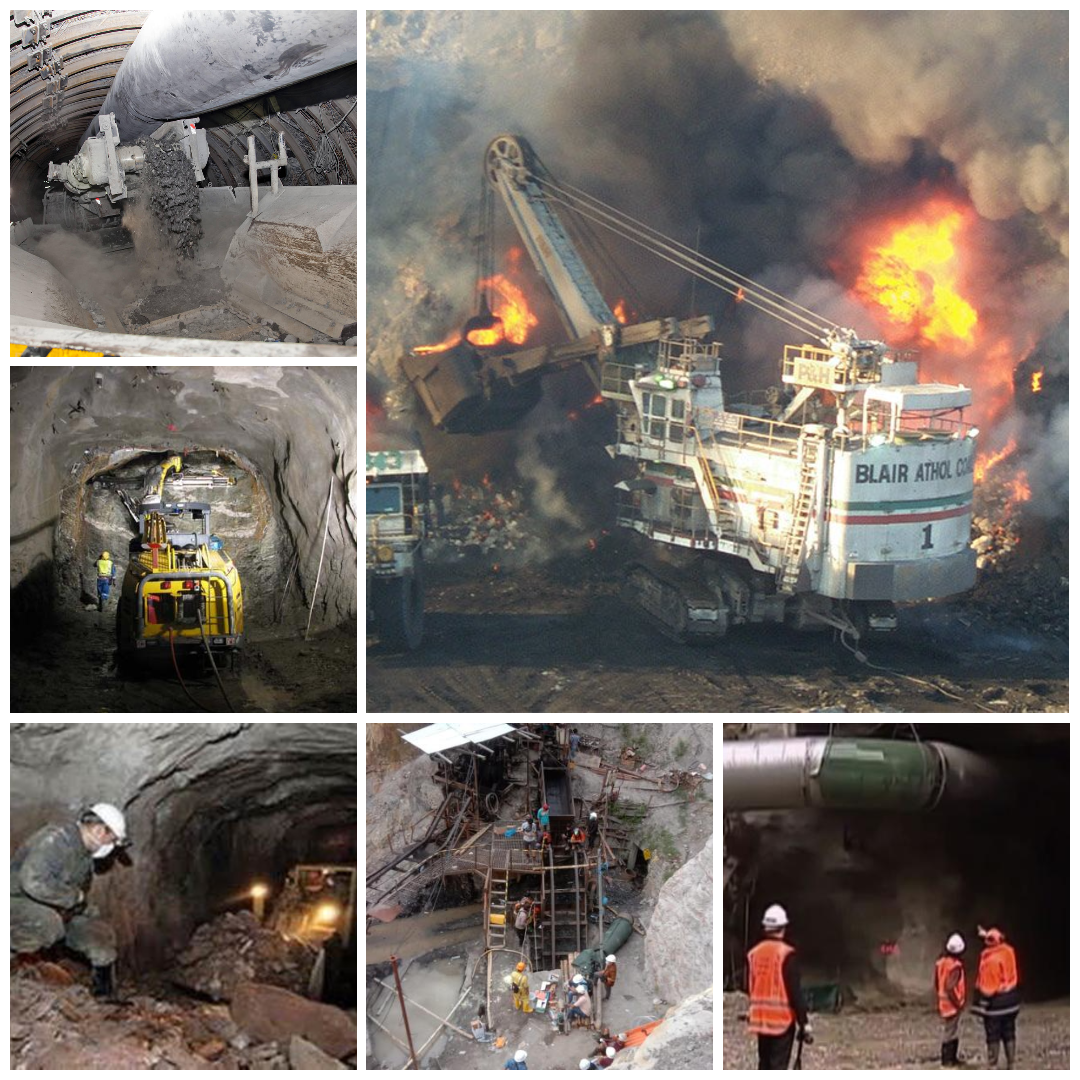} 
    \caption{Representative scenes depicting disaster scenarios in underground mining environments.}

    \label{fig:um_disaster_scenarios} \vspace{-0.6cm}
\end{figure}

Underground mining disasters present uniquely harsh and complex conditions that severely hinder situational awareness~\cite{jewel2024dis} 
and locating trapped personnel and coordinating effective rescue operations. However, conventional monitoring and communication systems often become inoperable due to structural collapse, power failure, or signal disruption, leaving responders with partial or outdated information (see Fig.\ref{fig:um_disaster_scenarios}). Existing situational awareness tools, typically based on discrete sensor inputs such as gas detectors or seismic monitors, offer limited insight and lack the contextual understanding necessary for a holistic scene interpretation \cite{yadav2025predicting}. Responders must often navigate in ``near-zero'' visibility caused by thick smoke, airborne dust, and total darkness, limiting their ability to detect hazards, locate victims, or understand the structural layout \cite{onifade2021towards}.

While image captioning has seen significant progress through models trained on large-scale general-purpose datasets such as the COCO caption dataset~\cite{chen2015microsoft} and Flickr30k~\cite{plummer2015flickr30k}, as well as more recent work focused on disaster imagery~\cite{weber2022incidents1m}, these approaches fall short when applied to the extreme and domain-specific conditions of underground mine disasters~\cite{jewel2024dis,goyal2025ogle}. Subterranean environments present unique vision challenges: narrow, confined spaces; structurally unstable scenes; low-light to dark conditions; high particulate density; and secondary hazards such as gas leaks and flooding. These conditions not only degrade image quality but also obscure critical visual cues necessary for accurate scene understanding and caption generation. Despite their potential, no existing vision-language framework has been designed to operate under such multimodal degradation and risk. This gap underscores the urgent need for {\em a robust framework capable of integrating noisy or degraded visual data with natural language reasoning} to generate context-aware, real-time descriptions that support situational awareness and decision-making in underground disaster response.

To this end, we propose the \textbf{Multimodal Disaster Situation Explainer (MDSE)}—the first vision-language framework 
to enhance post-disaster situational awareness in underground mines by integrating computer vision and natural language processing to automatically generate descriptive, actionable insights from real-time visual data captured in subterranean scenes. Trained on mining-domain data, MDSE detects critical hazards, structural damage, and potential escape routes, even under severe visual degradation. It incorporates mining-specific terminology and aligns with safety protocols to produce accurate, contextually relevant narratives interpretable by emergency responders. MDSE further introduces a novel contrast-aware cross-attention mechanism that improves feature fusion in low-light, smoke-filled, or occluded environments, enabling the system to extract subtle cues, such as cracked beams or blocked passages, and translate them into concise textual reports. In summary, the key contributions are as follows:

\begin{enumerate}[leftmargin=*]
    \item {\bf Novel Domain-Specific Captioning Framework:} We propose the first vision-language framework for mining disaster, introducing a \textit{context-aware cross-attention mechanism} that enhances salient visual features—such as edges and contours—in low-light, high-noise imagery, thereby improving captioning accuracy. 

    \item {\bf Vision Feature Enhancement and Segmentation-Aware Dual-Pathway Visual Encoding:} We propose a strategy that extracts both global scene-level features and object-centric regional features using a combination of a vision encoder and the Segment Anything Model (SAM) \cite{kirillov2023segment}. SAM produces precise segmentation masks even in dark, degraded images, isolating critical regions of interest. This allows the model to concentrate on key objects despite noise or occlusion in underground scenes. The dual-pathway design captures broad visual context while emphasizing salient regions, enhancing robustness in visually ambiguous conditions. 

    \item {\bf Integration of Low-Resource LLMs:} Instead of relying on an extremely large language model for caption generation, MDSE utilizes Parameter-Efficient Fine-Tuning with Low-Rank Adaptation (LoRA) on the Q-Former layers and LLM head, significantly reducing the number of trainable parameters during adaptation to downstream tasks, retaining the expressiveness of the original model while ensuring computational and memory efficiency.

\item {\bf Underground Mine Disaster (UMD) Dataset :} We construct the first image-captioning dataset specifically for underground mining emergencies, addressing a critical gap in vision-language research for extreme, low-visibility environments. To evaluate the effectiveness of our proposed framework, MDSE, we conduct comprehensive experiments on the UMD dataset, as well as on widely used benchmarks (COCO caption dataset~\cite{chen2015microsoft}, Flickr30k~\cite{plummer2015flickr30k}) and disaster-focused datasets (Incidents1M~\cite{weber2022incidents1m}, DNICC19k~\cite{zhou2023spatial}). Comparative analysis against state-of-the-art captioning models demonstrates that MDSE delivers robust and contextually accurate performance across both general and domain-specific scenarios, particularly excelling in the challenging conditions of underground disaster scenes.

\end{enumerate}

%% file: sec/2_related_work.tex
\section{Related Work} \label{sec:rel_work}

\noindent{\bf Vision-Language Models and Image Captioning.} The task of generating descriptive sentences from visual input has advanced significantly with the advent of deep learning, especially transformer-based architectures and vision-language pretraining \cite{xue2024xgen,agrawal2024pixtral,dai2023instructblip}. CLIP~\cite{radford2021learning} pioneered the alignment of textual and visual embeddings in a shared latent space, enabling strong {\em zero-shot} performance across multimodal tasks. BLIP~\cite{li2022blip} and its successor BLIP-2~\cite{li2023blip} further improved caption quality by incorporating multimodal encoders and decoders. Similarly, Flamingo~\cite{alayrac2022flamingo} offered a frozen vision-language transformer that set new benchmarks in few-shot and open-ended captioning tasks.

\noindent{\bf Specialized and Instruction-Tuned Models.} 
Detective networks~\cite{thanyawet2025detective} introduced attention-shifting mechanisms to enhance disaster image recognition, showing promise for improving contextual understanding in visually complex environments. LLaVA~\cite{liu2023visual} demonstrated the utility of instruction tuning for aligning vision-language models with user intent, boosting generalization in unseen domains. CrisisViT~\cite{long2024crisisvit} leveraged Vision Transformers with adversarial training on Incidents1M~\cite{weber2022incidents1m} and CrisisMMD~\cite{crisismmd2018icwsm}, achieving state-of-the-art performance across several disaster-related visual classification tasks.

\noindent{\bf Captioning Under Degraded Visual Conditions.} Captioning in low-light environments poses unique challenges due to diminished visual cues and increased noise. While research on captioning in such conditions is limited, related work in image enhancement, denoising, and low-light reconstruction has shown effectiveness in restoring perceptual quality. Techniques like  (SAM)~\cite{kirillov2023segment} offer robust segmentation capabilities, which can be integrated into captioning pipelines to help isolate critical scene components, even in dark or obscured settings.

\noindent{\bf Limitations in Underground Disaster Captioning.} Existing image captioning models have not addressed the unique challenges of domain-specific tasks such as disasters in underground mining, where extreme visual degradation and environmental hazards render standard approaches ineffective. This work fills that gap by proposing a tailored vision-language framework to enhance situational awareness and support decision-making in subterranean emergencies.

%% file: sec/3_methodology.tex
\begin{figure*}[!t] 
    \centering 
    \includegraphics[width=0.8\linewidth]{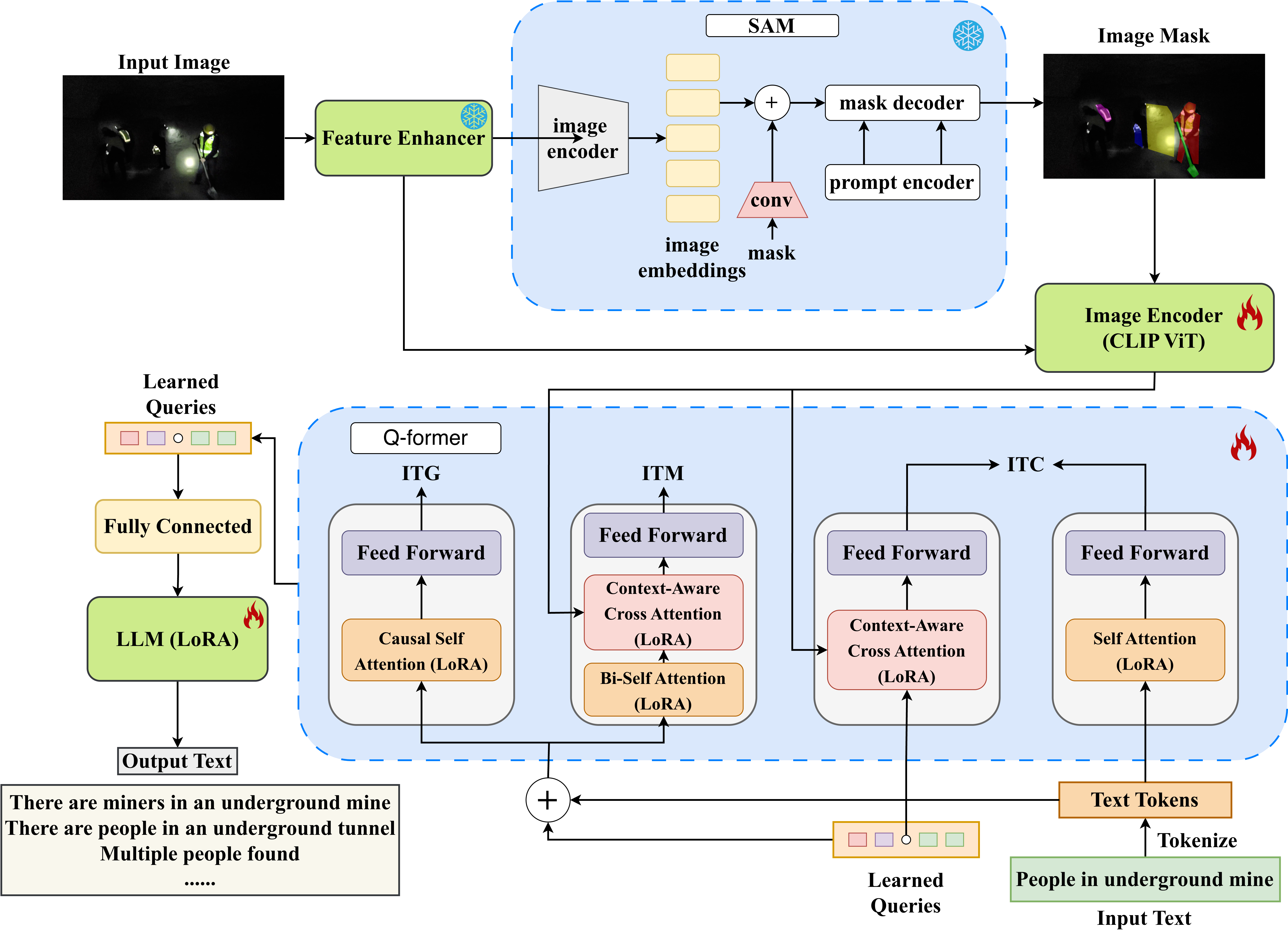}
     \caption{Overview of the proposed MDSE framework and its main components.}
    \label{fig:main_framework} 
\end{figure*}

\section{Methodology} \label{sec:method}
This section presents our proposed MDSE (Multimodal Disaster Situation Explainer) framework shown in \cref{fig:main_framework}.

\subsection{Feature Enhancement and Segmentation-Aware Dual-Pathway Encoding} 
Given an input image \(I\), a feature enhancement module \(\mathcal{F}_E\) first compensates for low light and noise by producing an enhanced image as $\hat{I} = \mathcal{F}_E(I)$. The feature enhancement module is designed as a sequential pipeline that improves image quality through several established operations. First, multi-exposure fusion combines differently exposed inputs using illumination-aware weighting to produce a well-balanced base image~\cite{vonikakis2011fusion}. Next, spatial tone mapping and local contrast enhancement boost detail visibility by center-surround filtering and multiscale contrast adjustment~\cite{vonikakis2008fast, vonikakis2016center}. White balance is then corrected using a combination of Gray-World and White-Point assumptions with a power transformation to remove color cast~\cite{vonikakis2011wb}. Finally, chromatic saturation is adaptively scaled to restore vividness and perceptual naturalness, yielding the enhanced output image. This ensures the vision encoder receives improved feature representations.

A pretrained SAM \(\mathcal{S}\) then generates segmentation masks $M$ highlighting key regions from the enhanced image $\hat{I}$ as $M = \mathcal{S}(\hat{I})$. The segmentation mask guides the model to focus on relevant objects or regions, reducing background noise during feature extraction. Even under poor visibility (darkness or smoke), SAM can delineate key objects and structures (e.g., equipment, hazards, tunnel walls) by generating multiple region-of-interest (ROI) masks in a single image. This allows  MDSE to isolate numerous relevant sub-scenes within the disaster image for focused analysis. The enhanced image \(\hat{I}\) and segmentation mask \(M\) are independently encoded by a CLIP ViT–based vision transformer \(\mathcal{V}_E\)~\cite{radford2021learning} as:\vspace{-0.1cm}
\begin{equation}
E_G = \mathcal{V}_E(\hat{I}), 
\qquad
E_S = \mathcal{V}_E(M).\vspace{-0.1cm}
\end{equation}
This dual‐pathway extraction produces robust visual embeddings at two scales: \(E_G\) captures the global scene embedding, while \(E_S\) encodes fine‐grained region-specific features within segmented sub‐regions. By integrating these complementary representations, MDSE maintains broad situational awareness and preserves critical localized evidence in challenging, low‐visibility environments.

Next, we fuse the global and segmented embeddings using a learnable projection matrix \(W_f\) as:
\begin{equation}
E_F = W_f \,[E_G; E_S],
\end{equation}
where \([E_G; E_S]\) denotes the concatenation of the global and segment-level embeddings. This fusion operation integrates complementary information from both pathways, yielding a rich and comprehensive feature representation.

MDSE processes both global and region-specific views to ensure that subtle peripheral details are retained. SAM-guided masks combined with CLIP encodings yield a multi-scale representation: the global embedding captures overall context, while the region-level embeddings preserve localized evidence— like a cracked beam or a miner’s helmet lamp—even under extreme low-visibility conditions.

\subsection{Context-Aware Cross-Attention Fusion}

Once robust global and region-specific embeddings are obtained, MDSE fuses them with linguistic context via a contrast-aware cross-attention module built on a BLIP-2  Q-Former \cite{li2023blip}. The Q-Former comprises two transformer submodules: i) {\em An image transformer} that refines and projects visual embeddings; and ii) {\em A text transformer} that encodes prompt tokens and autoregressively decodes captions. This design allows the model to dynamically focus on complementary visual features and generate coherent, context-aware descriptions.

External context $C$ is transformed by a learnable projection $f_{\text{ctx}}$ as
$C' = f_{\text{ctx}}(C)$ that integrates auxiliary context—such as scene type or task-specific cues—directly into the attention keys, enabling the model to prioritize visual features most relevant to the downstream task. Given the fused visual embedding \(E_F\) and context vector \(C'\), the context-enhanced key embeddings \(K'\) is computed as:
\begin{equation}
K' = E_F + \sigma\bigl(W_g\,[E_F; C']\bigr),
\end{equation}
where $E_F$ denotes final feature embeddings, $\sigma$ is the sigmoid activation,  and $W_g$ is a learnable gating projection. This gated fusion controls the degree of context influence on the visual features, allowing the model to modulate between relying on raw visual features and contextually enriched representations based on the task requirements. 

A fixed set of $N$ learnable query embeddings is provided as input to the image transformer (the number of queries and other Q-Former hyperparameters are specified in Sec.~\ref{sec:exp}). These query embeddings interact with each other through self-attention layers and engage with image features via context-aware cross-attention layers, which are inserted at regular intervals throughout the transformer blocks. Next, given queries $Q$, the attention computation is computed  as:
\begin{equation}
\text{Attention}(Q, K', V') = \text{softmax}\left( \frac{Q{K'}^T}{\sqrt{d_k}} \right) V'
\end{equation}
This operation enables dynamic interaction between query tokens and context-enriched visual embeddings, refining the attention focus based on the provided context. Furthermore, the queries can interact with text tokens through shared self-attention layers. Unlike the standard BLIP-2 implementation, where Q-Former uses fixed cross-attention with vision features, MDSE extends this by integrating a learned context signal directly into the key and value representations. This allows the cross-attention layers to adapt to both the visual input and the contextual cues, making the model more flexible and better suited for complex vision-language tasks, especially in diverse or challenging environments.

\subsection{Loss Function}
Following the Q-Former training paradigm, we jointly optimize a weighted sum of three loss terms:
\begin{equation}
\mathcal{L}_{\text{total}} = \lambda_{\text{1}} \cdot \mathcal{L}_{\text{ITC}} + \lambda_{\text{2}} \cdot \mathcal{L}_{\text{ITM}} + \lambda_{\text{3}} \cdot \mathcal{L}_{\text{ITG}}
\end{equation}
where \(\lambda_i\) is a weighting coefficient that balances the objectives, preventing any single task from dominating the learning process. The values of \(\lambda_i\) are tuned empirically to optimize performance across retrieval, matching, and captioning tasks. The individual loss terms are defined as follows: 
\begin{align}
{\mathcal{L}_{\mathrm{ITC}}}
&= \Resize{6.15cm}{-\frac{1}{N}\sum_{i=1}^{N}\sum_{j=1}^{N}
y_{ij}\log\frac{\exp\bigl(\mathrm{sim}(f_I^{i},f_T^{j})/\tau\bigr)}
{\sum_{k=1}^{N}\exp\bigl(\mathrm{sim}(f_I^{i},f_T^{k})/\tau\bigr)}},\\[6pt]
\mathcal{L}_{\mathrm{ITM}}
&= \Resize{6.15cm}{-\frac{1}{N} \sum_{i=1}^{N} \sum_{c=1}^{C} y_{i,c} \log \left( \frac{ \exp(\mathrm{ITM\_Logits}_{i,c}) }{ \sum_{k=1}^{C} \exp(\mathrm{ITM\_Logits}_{i,k}) } \right)},\\[6pt]
\mathcal{L}_{\mathrm{ITG}}
&= -\sum_{t=1}^{T} \log P\bigl(w_t \mid w_{<t},\,E_F\bigr).
\end{align}
Here \(\mathcal{L}_{\mathrm{ITC}}\) denotes the \emph{Image–Text Contrastive} loss, which aligns corresponding image and text features \(f_I\) and \(f_T\). The function \(\mathrm{sim}(\cdot,\cdot)\) computes the cosine similarity between visual embedding \(v_i\) and text embedding \(t_i\), scaled by a temperature \(\tau\). This objective encourages matching image–text pairs to have high similarity while pushing apart mismatched pairs. \(\mathcal{L}_{\mathrm{ITM}}\) denotes the \emph{Image–Text Matching} loss. The \(\mathrm{ITM\_Logits}\) are obtained from a linear classification head applied to the averaged query token embeddings produced by the image transformer after attending to the fused visual and textual features. This loss fine-tunes the model’s ability to distinguish between matched and mismatched image–text pairs, thereby directly enhancing performance on retrieval and matching tasks. \(\mathcal{L}_{\text{ITG}}\) denotes the \emph{Image-grounded Text Generation} loss, which corresponds to the autoregressive language modeling objective (i.e., captioning loss) over a sequence of length \(T\). At each time step \(t\), the model predicts the next word \(w_t\) conditioned on the previously generated tokens \(w_{<t}\) and the fused visual features \(E_F\). The probability \(P(w_t \mid w_{<t}, E_F)\) is computed by the autoregressive decoder. This loss guides the model to generate fluent, semantically accurate, and contextually grounded captions, word by word. The above combined objectives ensure cross-modal alignment, robust language understanding, and accurate caption generation. 

\subsection{Low-Rank Adaptation for LLM} 
For a model weight $W$, LoRA~\cite{hu2022lora} introduces low-rank matrices $A$ and $B$:
\begin{equation}
W' = W + A \cdot B
\end{equation}
with $A \in \mathbb{R}^{d \times r}$, $B \in \mathbb{R}^{r \times d}$, where $r \ll d$. LoRA allows efficient fine-tuning by updating only the low-rank factors, reducing computational overhead and memory usage. The parameter $d$ denotes the original weight dimension, $r$ is the rank deciding the compression factor, and $W'$ is the adapted weight matrix after low-rank modification. This method retains the expressiveness of the full model while enabling fast adaptation to new tasks with minimal parameters.  

In our MDSE framework, LoRA modules are injected into the linear projections of the Q-Former’s attention layers (i.e., the query, key, and value transformations). Specifically, for each projection matrix $W_Q, W_K, W_V$ in the self-attention and cross-attention layers, LoRA augments it as:
\begin{equation}
W_Q' = W_Q + A_Q \cdot B_Q, \newline
\end{equation}
\begin{equation}
W_K' = W_K + A_K \cdot B_K, \newline 
\end{equation}
\begin{equation}
W_V' = W_V + A_V \cdot B_V
\end{equation}

This design allows the Q-Former to adapt its attention dynamics to domain-specific scenarios (e.g., low-light disaster imagery) without altering the pretrained parameters. Since the Q-Former is the primary trainable component in BLIP-2, applying LoRA here yields efficient task-specific adaptation with minimal computational and memory overhead. On the LLM side, a LoRA adapter is placed on the projection that maps Q-Former outputs into the language model. We use LoRA rank 8, scale 16 for all Q-Former attention and MLP projections, and rank 4, scale 8 for the LLM input projector, with LoRA-dropout 0.05. 
In inference, the model first enhances and segments the input image, then fuses the resulting embeddings and applies context-aware cross-attention. The Q-Former and the LoRA-adapted LLM generate text autoregressively, enabling efficient multimodal reasoning in complex visual scenarios. This streamlined process ensures robust performance while maintaining computational efficiency.

%% file: sec/4_experiment.tex
\section{Performance Evaluation}
 \label{sec:exp}
\subsection{Experimental Setup}

We do not train SAM within MDSE; instead, SAM is used as an offline tool to generate region proposals that localize salient structures. The only visual backbone is CLIP ViT-L/14, which processes \(224\times224\) inputs with a \(14\times14\) patch size and outputs a sequence of \(257\) visual tokens of dimension \(1024\) (including the class token). The trainable part is the Q-Former(plus a small projector into the language model) with LoRA adapters. For query representation, we use $N = 32$ learnable queries of dimension $d = 768$, yielding \( \mathbf{Z}\in\mathbb{R}^{32\times768} \). The visual tokens are linearly projected from 1024 to 768 before cross-attention, and the Q-Former outputs are then projected to the language-model hidden size \(d_L=2560\) (OPT-2.7B)~\cite{zhang2022opt}. With LoRA rank \(r=8\) and scale \(\alpha=16\) on all Q-Former attention/MLP projections and on the LLM head, the total number of trainable parameters is \(\approx 2.0\)M. The loss weights \((\lambda_{\text{ITC}},\lambda_{\text{ITM}},\lambda_{\text{ITG}})\) were tuned via a small grid on the validation split where we adopt \((1.0,\,0.5,\,0.10)\) based on the best performance.

\noindent{\bf Datasets.}  To evaluate MDSE against state-of-the-art approaches, we use several benchmark datasets for image captioning: Flickr30k~\cite{plummer2015flickr30k}, COCO Captions~\cite{chen2015microsoft}, Incidents1M~\cite{weber2022incidents1m}, CDNIC19k~\cite{zhou2023spatial}, and our UMD dataset. Brief descriptions of these datasets are provided below:
\begin{figure}[!t] 
\centering 
\includegraphics[width=0.9\linewidth]{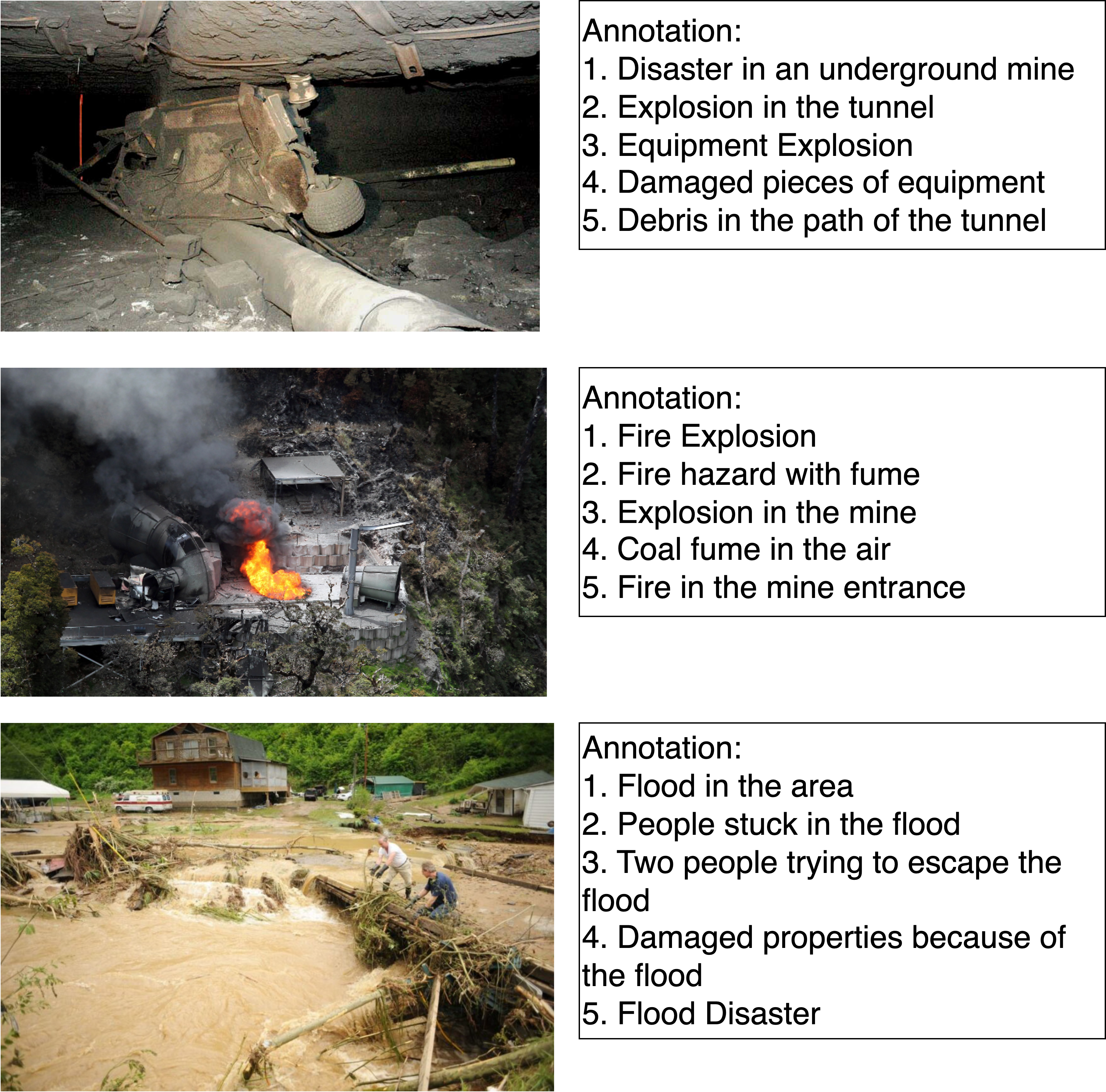} 
\caption{Representative samples from the UMD dataset with corresponding textual annotations.}
\label{fig:um_datset} \vspace{-6mm}
\end{figure}

\begin{itemize}
    \item{\bf Underground Mine Disaster (UMD) Dataset:} We introduce the (UMD) dataset, a domain-specific image–caption corpus designed to advance visual recognition and situational understanding in underground mining disaster scenarios, an area largely absent from existing computer vision benchmarks. Such events present unique challenges, including confined spaces, poor lighting, smoke, debris, and structural collapse, which require specialized models for reliable detection and assessment. To address the scarcity of publicly available imagery, we curated the dataset through keyword-based internet searches and supplemented it with visually realistic scenes from disaster films that simulate underground emergencies. These cinematic samples provide diverse, contextually rich depictions of hazards, rescue efforts, and structural damage. UMD contains 300 manually annotated images with five captions for each image, partially drawn from Image-Mine~\cite{jewel2024dis}, DNICC19k~\cite{zhou2023spatial}, and Incidents1M~\cite{weber2022incidents1m}. Representative examples are shown in \cref{fig:um_datset}. To standardize heterogeneous sources, all images are de-duplicated, lightly border-trimmed, color-normalized, and resized to $224 \times 224$ pixels for both training and inference. 

    \item {\bf Flickr30k~\cite{plummer2015flickr30k}:} It is a widely used image captioning dataset with roughly 31,783 images from Flickr, each annotated with five human-written English captions. Covering diverse everyday scenes, it serves as a standard benchmark for evaluating captioning models.

    \item {\bf COCO caption dataset~\cite{chen2015microsoft}:} Microsoft COCO (Common Objects in Context) captions is a large-scale dataset comprising over 200,000 images annotated for tasks such as object detection, segmentation, and image captioning. Each image is paired with five captions, providing strong language grounding for vision-language modeling.

    \item {\bf Incidents1M~\cite{weber2022incidents1m}:} A large-scale dataset with one million images of real-world disasters, accidents, and critical incidents, including floods, earthquakes, fires, and infrastructure failures. It provides annotations for incident type, damage extent, and context, enabling tasks such as disaster classification and automated damage assessment. Its various event types, regions, and conditions make it a challenging benchmark for scene understanding and advanced research in crisis response and emergency control.

    \item {\bf DNICC19k~\cite{zhou2023spatial}:}  It is known as CDNIC19k, a disaster-focused dataset with $\sim$19,000 images across 15 categories, including natural hazards, accidents, and emergencies. It provides annotations for damage assessment, incident classification, and captioning. While less widely used than COCO or Flickr30k, DNICC19k offers event-specific imagery with contextual labels, supporting research in disaster control and humanitarian applications.
\end{itemize}

\noindent {\bf Baselines.} For benchmarking, we consider a set of strong baseline vision–language models that allow a comprehensive assessment of MDSE’s effectiveness:
\begin{itemize} 
    \item {\bf BLIP-2~\cite{li2023blip}:} A vision-language model using a frozen image encoder and a learnable Q-Former to bridge vision and language. Excels in multimodal tasks like captioning and visual question answering.
    
    \item {\bf Flamingo~\cite{alayrac2022flamingo}:} DeepMind’s few-shot multimodal model using gated cross-attention with a frozen image encoder and a language model for flexible multimodal tasks.
    
    \item {\bf CLIP~\cite{radford2021learning}}— A contrastive vision-language model aligning images and texts for robust zero-shot recognition and retrieval, widely used for visual understanding. 
    
    \item {\bf LLaVA~\cite{liu2023visual}:} A large vision-language assistant augmenting a Vicuna LLM with visual grounding through lightweight projection and instruction-tuning on multimodal dialogues.
    
    \item {\bf Florence-2~\cite{xiao2024florence}:} Microsoft’s unified vision-language transformer model trained for tasks like captioning, detection, and grounding on large-scale datasets. 

    \item {\bf Florence~\cite{yuan2021florence}:} Florence is a large-scale vision foundation model designed to learn generalizable visual representations from diverse datasets, serving as a strong baseline for tasks such as image classification, object detection, and image captioning.

    \item {\bf STCNet~\cite{zhou2023spatial}:} A disaster-domain captioner that leverages spatial context and topic guidance; it surpasses Show-Attend-and-Tell~\cite{xu2015show} and AoANet~\cite{gao2022aonet} on DNICC19k, making it a strong baseline for MDSE.
 
    \item {\bf Qwen-2.5~\cite{yang2025qwen3}:} Excels at integrating visual and textual features, producing accurate, detailed image descriptions through cross-modal reasoning.
\end{itemize}

\noindent {\bf Evaluation metrics.} For performance comparison on the UMD and DNICC19k datasets, we use Consensus-based Image Description Evaluation (CIDEr)~\cite{vedantam2015cider} and Semantic Propositional Image Caption Evaluation (SPICE)~\cite{anderson2016spice} scores. Below is a brief description of each one:
\begin{itemize}
    \item {\bf CIDEr~\cite{vedantam2015cider}:} It measures similarity between a generated caption and human references using TF-IDF weighted \(n\)-gram matching, rewarding both distinctive content (via higher weights for rare \(n\)-grams) and commonly agreed phrases, yielding a robust measure of image relevance.

    \item {\bf SPICE~\cite{anderson2016spice}:} It evaluates by comparing their semantic scene graphs—objects, attributes, and relationships—between generated and reference captions. It computes an F-score over matching tuples, focusing on meaning rather than surface \(n\)-gram overlap and aligning more closely with human judgment.

\end{itemize}
For the COCO caption and  Flickr30k datasets, we evaluate models using recall-based metrics.  In our experimental settings, recall at rank \(k\) (R@k) measures the proportion of queries for which the correct caption appears within the top \(k\) predictions.  Formally, let \(N\) be the total number of queries, and let \(r_i\) be the rank at which the correct label appears for the \(i\)-th query. Then R@k is defined as:\vspace{-0.1cm}
\[
\text{R@k} = \frac{1}{N}\sum_{i=1}^{N} \mathbb{I}(r_i \leq k)\vspace{-0.1cm}
\]
where $\mathbb{I}$ is the indicator function, which returns 1 when the condition $r_i \leq k$ is true and 0 otherwise. In practice, R@1 measures the fraction of instances where the correct answer is the top prediction, while R@5 and R@10 evaluate how often the correct answer is found within the top 5 or top 10 predictions, respectively.

\begin{figure*}[!t]
    \centering
    \begin{subfigure}[t]{0.98\textwidth}
        \centering
        \includegraphics[width=\linewidth]{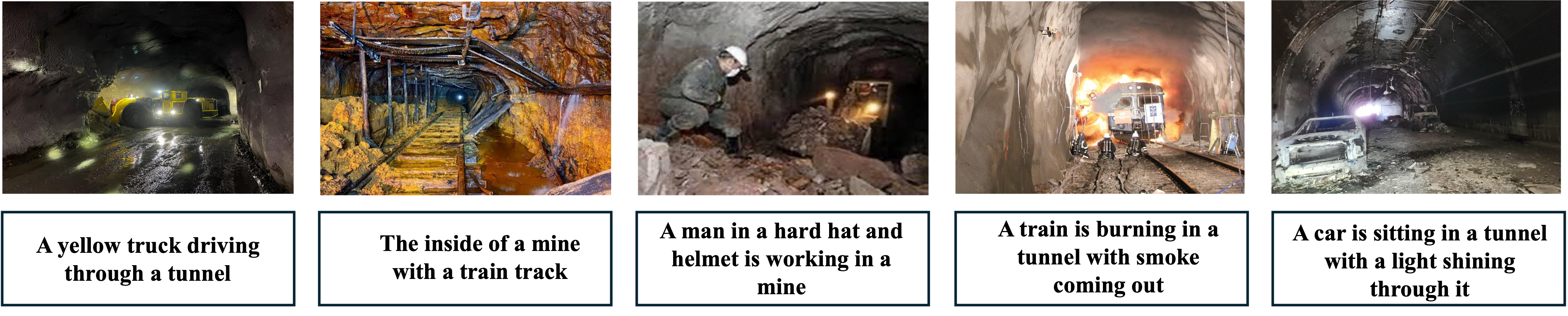}
\caption{Sample images from the UMD dataset}
        \label{fig:subfig1}\vspace{-1mm}
    \end{subfigure}

    \vspace{0.5em} 

    \begin{subfigure}[t]{0.98\textwidth}
        \centering
        \includegraphics[width=\linewidth]{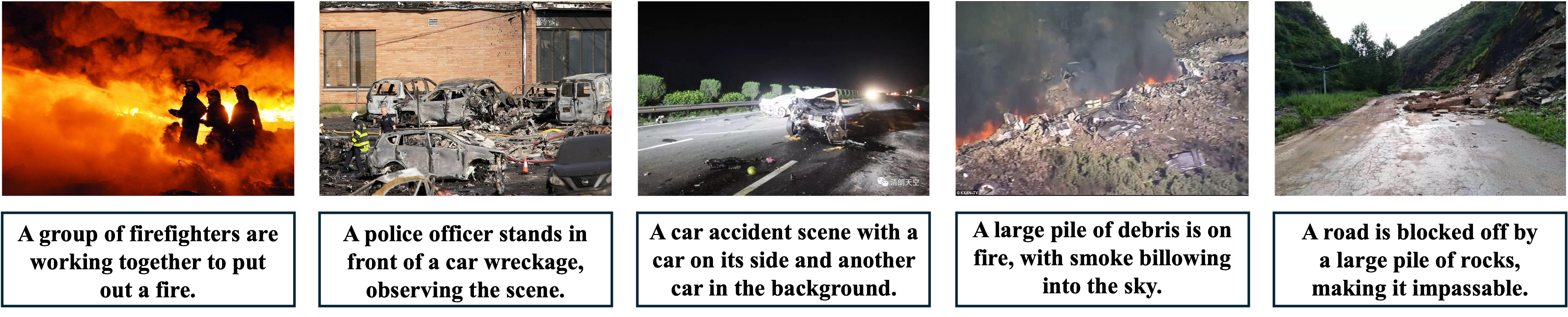}
\caption{Sample images from the DNICC19k dataset.}        \label{fig:subfig2}\vspace{-2mm}
    \end{subfigure}

\caption{MDSE inference results on UMDD and DNICC19k datasets, showing its captioning performance under challenging conditions.}
\vspace{-4mm}

    \label{fig:doublecolumn_stacked}
\end{figure*}


 Finally, for the Incidents1M dataset, we use the prompt {\em ``Describe the incidents depicted in this image among these labels''}, as the dataset provides class-level labels rather than sentence-level annotations. We use mean Average Precision (mAP) as the primary evaluation metric, which is widely employed in multi-label classification and retrieval tasks. For each class, the Average Precision (AP) is computed by plotting the precision-recall curve and calculating the area under the curve (AUC). The mAP is then obtained by averaging the AP values across all classes. This metric effectively captures both precision and recall, making it well-suited for handling class imbalance and multi-label scenarios, as present in the Incidents1M dataset. 

\subsection{Experimental Results}



We evaluate the proposed \text{MDSE} framework against baselines using \text{CIDEr} and \text{SPICE} on the \text{UMD} and \text{DNICC19k} datasets. To test generalization, we compare performance on standard benchmarks (\text{MSCOCO} and \text{Flickr30k}) using image-text retrieval metrics. We also assess classification on the disaster-oriented \text{Incidents1M} dataset via zero-shot evaluation with a fixed prompt. All models are fine-tuned on UMD, DNICC19k, COCO, and Flickr30k.

\begin{table}[!t]
\centering
\caption{Performance comparison of MDSE with baseline image captioning methods.}
\setlength{\tabcolsep}{6pt}
\renewcommand{\arraystretch}{ 0.9}
\label{tab:image-captioning-comp}
\resizebox{\linewidth}{!}{
\begin{tabular}{lcccc}
\toprule
\textbf{Model} & \multicolumn{2}{c}{\textbf{UMD}} & \multicolumn{2}{c}{\textbf{DNICC19k}} \\
\cmidrule(lr){2-3} \cmidrule(lr){4-5}
& \textbf{CIDEr} & \textbf{SPICE} & \textbf{CIDEr} & \textbf{SPICE} \\
\midrule
BLIP-2     & 0.62 & 0.50 & 0.58 & 0.45 \\
Flamingo   & 0.53 & 0.42 & 0.53 & 0.41 \\
Florence   & 0.58 & 0.47 & 0.57 & 0.43 \\
LLaVA      & 0.59 & 0.47 & 0.57 & 0.43 \\ 
STCNet     &  --  &  --  & 0.53 & 0.43 \\
Qwen-2.5     &  0.66  &  0.51  & 0.62 & 0.47 \\
MDSE (Ours) & 0.70 & 0.53 & 0.66 & 0.49 \\
\bottomrule
\end{tabular}
}\vspace{-5mm}
\end{table}

\Cref{tab:image-captioning-comp} presents a comparison of our proposed MDSE method with baseline image captioning models on two disaster-related datasets: UMD and DNICC19k~\cite{zhou2023spatial}. On the UMD dataset, MDSE significantly outperforms all other models. It achieves a CIDEr score of 0.70, which is notably higher than the next best model, BLIP-2, which scored 0.62. MDSE also surpasses other baselines such as LLaVA, Qwen, and Florence. For the SPICE metric, MDSE scores 0.53, exceeding BLIP-2's score of 0.50 and showing clear improvements over the other methods. These results demonstrate MDSE’s strong capability to produce captions that are both syntactically coherent and semantically meaningful. On the DNICC19k dataset, MDSE again shows superior performance, achieving a CIDEr score of 0.66 and a SPICE score of 0.49, both higher than those of the competing models. Additionally, \Cref{fig:doublecolumn_stacked} highlights MDSE's consistent improvements across datasets and metrics, demonstrating its effectiveness in producing human-aligned image descriptions on real-world datasets.

In \Cref{tab:image-text-retrival-comp}, we show image-text retrieval results across COCO caption and Flickr30k datasets using \text{Recall at $k$ (R@1, R@5, R@10)}. These metrics reflect the percentage of queries for which a relevant item is retrieved within the top-$k$ results, where higher values indicate better retrieval accuracy. On the \text{COCO caption dataset}, MDSE outperforms all baselines. It achieves the highest R@1 score of \text{0.88}, surpassing \text{BLIP-2}, which  achieves \text{0.84}. For R@5, MDSE scores \text{0.97}, slightly above BLIP-2's \text{0.96}. In R@10, both MDSE and BLIP-2 attain a perfect \text{0.98}. Competing models like \text{Flamingo} and \text{Florence} lag behind, with R@1 scores of \text{0.66} and \text{0.68}, respectively. Similarly, on \text{Flickr30k}, MDSE also demonstrates strong performance, matching BLIP-2’s R@1 score of \text{0.98}. For both R@5 and R@10, MDSE achieves perfect scores of \text{1.00}, equaling \text{BLIP-2} and \text{Flamingo}. These results highlight MDSE’s ability to retrieve semantically relevant captions and its generalization across both domain-specific and general-purpose benchmarks.

\begin{table}[!t]
\centering
\caption{Comparison with baseline image-text retrieval methods.}
\label{tab:image-text-retrival-comp}
\resizebox{\linewidth}{!}{
\begin{tabular}{lcccccc}
\toprule
\textbf{Model} & \multicolumn{3}{c}{\textbf{COCO Caption Dataset}} & \multicolumn{3}{c}{\textbf{Flickr30k Dataset}} \\
\cmidrule(lr){2-4} \cmidrule(lr){5-7}
& \textbf{R@1} & \textbf{R@5} & \textbf{R@10} & \textbf{R@1} & \textbf{R@5} & \textbf{R@10} \\
\midrule
BLIP-2       & 0.84 & 0.96 & 0.98 & 0.98 & 1.00 & 1.00 \\
CLIP         & 0.60 & 0.83 & 0.90 & 0.88 & 0.99 & .99 \\
Flamingo     & 0.66 & 0.87 & 0.93 & 0.89 & 0.99 & 1.00 \\
Florence     & 0.68 & 0.86 & -- & 0.91 & .99 & -- \\
Qwen-2.5     & 0.87 & 0.97 & 0.98 & 0.99 & 1.00 & 1.00 \\
MDSE (Ours)  & 0.88 & 0.97 & 0.98 & 0.98 & 1.00 & 1.00 \\
\bottomrule
\end{tabular}}\vspace{-2mm}
\end{table}

\begin{table}[!t]
\centering
\caption{Zero-shot classification results on Incidents1M dataset.}

\label{tab:class_eval_metrics_incidents_1m}
\resizebox{\linewidth}{!}{
\begin{tabular}{lccccc}
\toprule
\textbf{Model} & BLIP-2  &CLIP &Flamingo&Florence-2&MDSE (Ours) \\
\midrule
   \textbf{mAP}& 0.44 &0.50&0.45&0.52&0.47\\
\bottomrule
\end{tabular}}\vspace{-1mm}
\end{table}

\begin{figure}[!t] 
\centering 
\includegraphics[width=\linewidth]{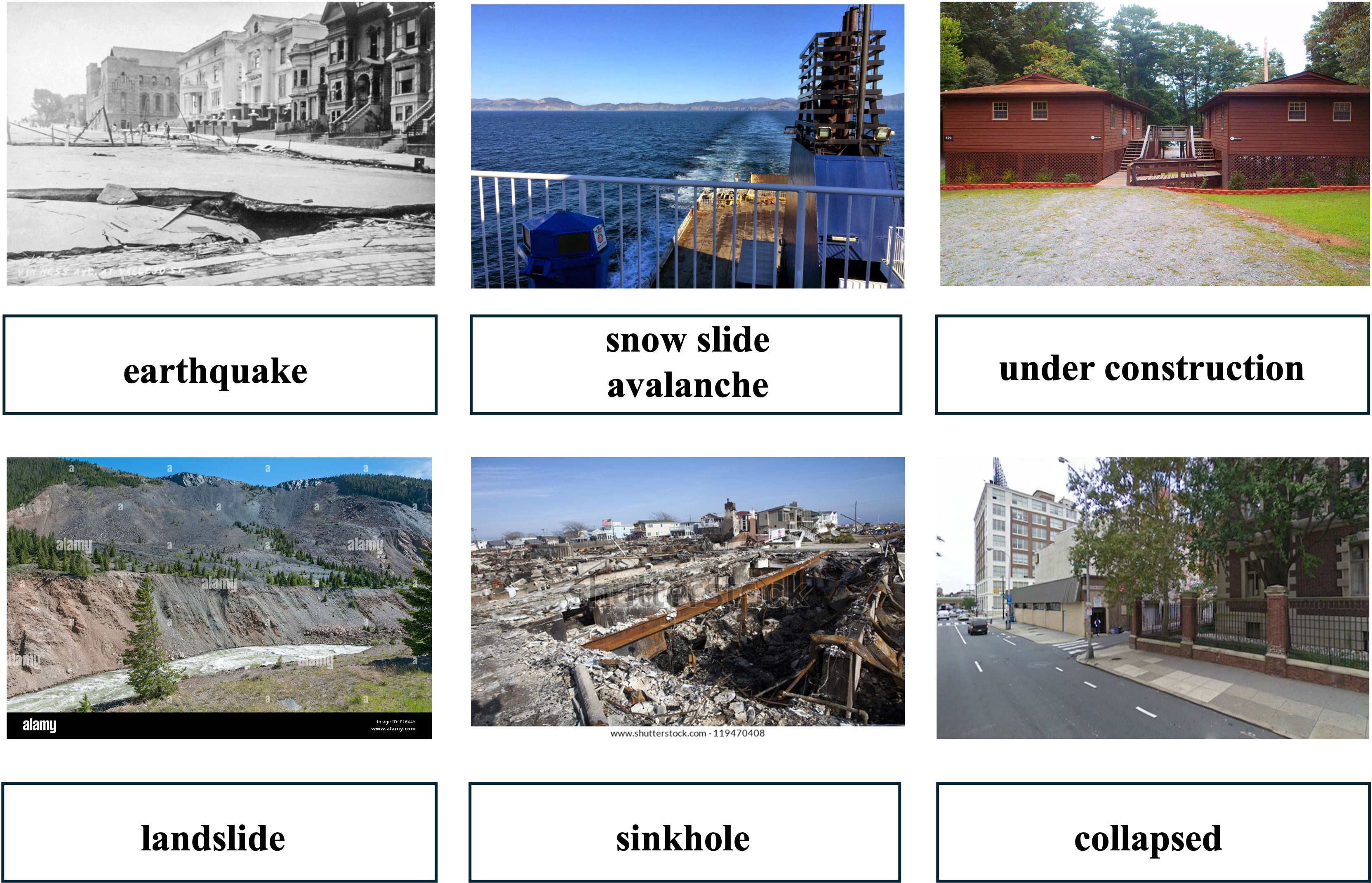} 
\caption{Demo results of MDSE on the Incidents1M dataset.}
\label{fig:demo_incident}\vspace{-1mm}
\end{figure}

Finally, in \Cref{tab:class_eval_metrics_incidents_1m}, we report the mAP scores of our MDSE approach compared to the baselines on the Incidents1M dataset for zero-shot classification. Among the baselines, Florence-2 achieves the highest mAP score of 0.52, followed by CLIP with 0.50. Our MDSE model achieves a competitive mAP of 0.47, outperforming BLIP-2 (0.44) and Flamingo (0.45), though it does not surpass the top-performing models. Additionally, \Cref{fig:demo_incident} presents qualitative examples of MDSE’s classification outputs. The model correctly classifies three images corresponding to earthquake, landslide, and sinkhole, while the remaining examples are misclassified. 

It is worth emphasizing that the primary objectives of MDSE are image captioning and image-text retrieval, where it has shown state-of-the-art performance, as detailed in earlier sections. In contrast, zero-shot classification is a core strength of models like CLIP and Florence-2, which are explicitly designed for broad visual recognition and multi-task vision-language understanding. Despite this distinction, MDSE demonstrates robust generalization to unseen incident categories and provides strong zero-shot classification performance on Incidents1M (see \Cref{fig:demo_incident}). These results highlight MDSE’s versatility and balanced performance across diverse vision-language tasks.

\subsection{Ablation Study} \label{sec:ablation}

We ablate MDSE by removing one component at a time and report CIDEr/SPICE on UMD and DNICC19k (\Cref{tab:ablation}). The largest degradations arise from removing dual-path fusion (global+region) and removing SAM-based region proposals (offline), confirming the value of combining global/region evidence and localized proposals. If we disable the Context-aware cross-attention and use the cross-attention module from the original Q-former, we find lower SPICE scores for both datasets, indicating weaker relational detail. Removing the feature enhancement module yields only a small decline, suggesting FE improves robustness but is not the dominant driver of gains.

\begin{table}[!t]
\centering
\caption{MDSE 's evaluation on UMD and DNICC19k datasets.}
\label{tab:ablation}
\resizebox{\linewidth}{!}{
\begin{tabular}{lcccc}
\toprule
\textbf{Model} & \multicolumn{2}{c}{\textbf{UMD}} & \multicolumn{2}{c}{\textbf{DNICC19k}} \\
\cmidrule(lr){2-3} \cmidrule(lr){4-5}
& \textbf{CIDEr} & \textbf{SPICE} & \textbf{CIDEr} & \textbf{SPICE} \\
\midrule
w/o Dual-Pathway Encoding     & 0.65 & 0.5 & 0.62 & 0.47 \\
w/o SAM   & 0.66 & 0.51 & 0.63 & 0.47 \\
w/o Context aware cross attention & 0.67 & 0.5 & 0.64 & 0.46 \\
w/o FE    & 0.68 & 0.53 & 0.64 & 0.48 \\ 
MDSE (whole) & 0.70 & 0.53 & 0.66 & 0.49 \\
\bottomrule
\end{tabular}
}
\end{table}
\begin{table}[!t]
\centering
\setlength{\tabcolsep}{6pt}
\caption{LoRA hyperparameter sweep on UMD and DNICC19k (r = rank, $\alpha$ = scale).}
\label{tab:lora_ablation}

\renewcommand{\arraystretch}{ .5}
\resizebox{6.5cm}{!}{
\begin{tabular}{lcccc}
\toprule
\textbf{(r, $\alpha$)} & \multicolumn{2}{c}{\textbf{UMD}} & \multicolumn{2}{c}{\textbf{DNICC19k}} \\
\cmidrule(lr){2-3} \cmidrule(lr){4-5}
& \textbf{CIDEr} & \textbf{SPICE} & \textbf{CIDEr} & \textbf{SPICE} \\
\midrule
(4,\;4)               & 0.68 & 0.52 & 0.64 & 0.48 \\
(4,\;8)               & 0.69 & 0.52 & 0.65 & 0.48 \\
(8,\;8)               & 0.69 & 0.53 & 0.65 & 0.49 \\
(8,\;16)              & 0.70 & 0.53 & 0.66 & 0.49 \\
(16,\;16)             & 0.69 & 0.52 & 0.65 & 0.48 \\
(16,\;32)             & 0.68 & 0.51 & 0.64 & 0.47 \\
\bottomrule
\end{tabular}
}\vspace{-3mm}
\end{table}

We also sweep LoRA hyperparameters ($r,\alpha$) on the Q-Former and LLM input projector \Cref {tab:lora_ablation}. Performance is stable across settings, with a mild peak in the middle on both datasets. These results indicate that modest-rank adapters provide efficient task adaptation without large compute/memory overhead, while more aggressive ranks do not yield additional gains under our data scale.

%% file: sec/6_conclusion.tex
\section{Conclusion}
This paper presents MDSE, a new captioning framework tailored for low-light and visually challenging underground mine disaster scenes. MDSE combines segmentation-aware dual-pathway visual encoding with context-aware cross-attention, enabling robust alignment of visual and textual data under severe degradation. Experiments on our newly introduced UMD dataset demonstrate that MDSE outperforms state-of-the-art baselines, generating more accurate and domain-appropriate descriptions, with a 12\% improvement over leading vision-language models.